\documentclass[default]{sn-jnl}

\usepackage[utf8]{inputenc}
\usepackage{graphicx}
\usepackage{array,booktabs}
\usepackage{tabularx}
\usepackage{lscape}
\usepackage{tabularray}
\usepackage{xcolor}
\usepackage{makecell}
\usepackage{multirow}
\usepackage{tikz}
\usetikzlibrary{arrows,shapes}
\usetikzlibrary{arrows.meta}
\usepackage{dsfont}
\usepackage{amsmath, amsthm, amssymb,amsfonts}
\usepackage{mathrsfs}%
\theoremstyle{definition}
\newtheorem{definition}{Definition}
\usepackage{enumitem}
\usepackage{float}
\usepackage{caption}
\usepackage{subcaption}
\usepackage{todonotes}
\usepackage{xcolor}
\usepackage[title]{appendix}%
\usepackage{textcomp}%
\usepackage{manyfoot}%
\usepackage{algorithm}%
\usepackage{algorithmicx}%
\usepackage{algpseudocode}%
\usepackage{listings}%
\usepackage{footnote}
\makesavenoteenv{tabular}
\makesavenoteenv{table}

\newcolumntype{s}{>{\arraybackslash}m{0.001cm}}
\newcommand{\pr}{\mathds{P}}
\def\krr{{$k$-RR}}

\begin{document}

\title{On the Impact of Multi-dimensional Local Differential Privacy on Fairness}

\author[1,3]{\fnm{Karima} \sur{Makhlouf}}

\author[2]{\fnm{Héber H.} \sur{Arcolezi}}

\author[3]{\fnm{Sami} \sur{Zhioua}}

\author[4]{\fnm{Ghassen} \sur{Ben Brahim}}
 
\author[1,3]{\fnm{Catuscia} \sur{Palamidessi}}

\affil[1]{\orgdiv{Inria}, \orgaddress{\city{Palaiseau}, \country{France}}}

\affil[2]{\orgdiv{Inria Centre at the University Grenoble Alpes}, \orgaddress{\city{Grenoble}, \country{France}}}

\affil[3]{\orgdiv{École Polytechnique (IPP)}, \orgaddress{\city{Palaiseau}, \country{France}}}

\affil[4]{\orgdiv{College of Computer Engineering and Science, Prince Mohammad Bin Fahd University}, \orgaddress{\city{Dammam}, \country{Saudi Arabia}}}

\abstract{

Automated decision systems are increasingly used to make consequential decisions on people's lives. 
Due to the sensitivity of the manipulated data as well as the resulting decisions, several ethical concerns need to be addressed for the appropriate use of such technologies, in particular, fairness and privacy. 
Unlike previous work which focused on centralized differential privacy (DP) or on local DP (LDP) for a single sensitive attribute, in this paper, we examine the impact of LDP in the presence of several sensitive attributes (\textit{i.e., multi-dimensional data}) on fairness.
Detailed empirical analysis on synthetic and benchmark datasets revealed very relevant observations. In particular, (1) multi-dimensional LDP is an efficient approach to reduce disparity, (2) the multi-dimensional approach of LDP (independent vs combined) matters only at low privacy guarantees (high $\epsilon$), and (3) the outcome $Y$ distribution has an important effect on which group is more sensitive to the obfuscation. 
Last, we summarize our findings in the form of recommendations to guide practitioners in adopting effective privacy-preserving practices while maintaining fairness and utility in ML applications.

}

\keywords{Differential Privacy, Machine learning, Fairness, Randomized Response}



\maketitle

\section{Introduction}
\label{intro}

Data collected about individuals is regularly used to make decisions that impact those same individuals. For example, census statistics have important implications for all aspects of daily life, including the allocation of political power, the distribution of federal funds, and research in economics and social sciences. In banking industries, machine learning (ML) models leverage data to proactively monitor customer behavior, reduce the likelihood of false positives, and prevent fraud.
In these settings, there is a tension between the need for accurate systems, in which individuals receive what they deserve, and the need to protect individuals from improper disclosure of their sensitive information. 
Differential privacy (DP)~\cite{dwork2016calibrating} is now widely recognized as the gold standard for providing formal guarantees on the privacy level achieved by an algorithm. However, central DP can only be used on the assumption of a trustworthy server. Local DP (LDP)~\cite{kasiviswanathan2011can} is a variant that achieves privacy guarantees for each user locally with no assumptions on third-party servers. 
In other words, LDP ensures that each user’s data is locally obfuscated first on the client-side and then sent to the server-side, thus protecting data from privacy leaks on both the client-side and the server-side. 
Many Big tech companies have deployed LDP-based algorithms to use in their industrial products (e.g., Google Chrome~\cite{erlingsson2014rappor} and Apple iOS~\cite{apple_2017}).

On the other hand, algorithmic fairness aims to ensure that induced models do not discriminate against groups or individuals based on their protected\footnote{In this paper, we use the term \textit{protected} to designate sensitive attributes from a fairness perspective and the term \textit{sensitive} to designate sensitive attributes from a privacy perspective.} attributes (e.g., race, gender, age, etc.). Several fairness notions have been formally defined and proposed in the literature in order to assess/quantify discrimination~\cite{10.1145/3468507.3468511}. These fairness notions fall into two main categories namely, group and individual notions. Group fairness notions aim to ensure that sub-populations have similar decisions while individual fairness notions aim to ensure that similar individuals are treated equally \cite{makhlouf2021Bridging, alves2022survey, mitchell2021algorithmic, 10.1145/3457607}.   

Striking a balance between privacy and fairness while maintaining utility is crucial. However, privacy-preserving algorithms in particular, DP, may tend to disparately affect members of minority groups, implying that privacy and fairness are fundamentally at odds~\cite{farrand2020neither, bagdasaryan2019differential, chang2021privacy, fioretto2022differential, ganev2022robin}. This tension between fairness and DP is attracting more and more attention, however, a clear understanding of the reasons for this tension is still not well explored. In another line of research, DP and fairness were viewed as aligned objectives. For instance, Dwork et al.~\cite{dwork2012fairness} proved that individual fairness is a generalization of DP and provided some constraints under which a DP mechanism ensures individual fairness as well. Alternatively, DP and fairness have been integrated as dual objectives in a learning model. For instance, Xu et al.~\cite{xu2019achieving} proposed two algorithms to achieve both DP and fairness in logistic regression by combining functional mechanism and decision boundary fairness. 

In this paper, we investigate the impact of training a model with obfuscated data under LDP guarantees, employing the well-known $k$-ary Randomized Response (\krr)~\cite{kairouz2016discrete} mechanism. The choice of \krr{} is motivated by its optimality for distribution estimation under several information theoretic utility functions~\cite{kairouz2016discrete} and also it's design simplicity since \krr{} does not require any particular encoding. Specifically, since the output space is equal to the input space, \krr{} provides optimal computational and communication costs for users. Moreover, on the server side, no decoding step is needed. It also means that the server is free to use any post-processing coding techniques (e.g., one-hot encoding, mean encoding, binary encoding) to improve the usefulness of the ML model.

\krr{} has traditionally been mainly employed in the one-dimensional scenario in LDP and fairness literature~\cite{mozannar2020fair,chen2022fairness}, where only one attribute is randomized. However, relying solely on LDP for a single sensitive attribute might be insufficient. This limitation stems from potential correlations that could allow attackers to reconstruct the privatized sensitive attribute. Hence, we specifically address scenarios involving multiple sensitive attributes, providing a more realistic representation of data collections in real-world contexts. Nevertheless, applying \krr{} to multi-dimensional sensitive data presents greater challenges~\cite{Domingo2022,kikuchi2022castell}. For example, the naive approach of obfuscating each sensitive attribute independently results in the loss of any dependencies between sensitive attributes. This method has been recently employed to evaluate the impact of LDP on fairness~\cite{Arcolezi2023}. In our study, in addition to this independent setting, we also explore a combined setting that merges all sensitive attributes into a single attribute. Indeed, combined \krr{} has not been extensively studied, and its impact on fairness remains unclear, a gap in understanding that we aim to address.

More specifically, the contributions of this paper are threefold. First, we study the impact of LDP on fairness and utility by observing the behaviour of sub-populations separately. This allows for a more complete understanding of how the fairness metrics behave under different LDP guarantees. Second, we compare both independent and combined settings for obfuscating multi-dimensional sensitive attributes under LDP guarantees. Third, we study how the target distribution has an impact on the privacy-fairness-utility trade-off. 
The key findings of our empirical analysis are:
\begin{enumerate}
\item Generally, obfuscating data with LDP contributes generally to reduce disparity.
\item Obfuscating several sensitive attributes (multi-dimensional) reduces disparity more efficiently than obfuscating a single attribute (one-dimensional).
\item The multi-dimensional approaches of LDP (independent vs combined) differ in their impact on fairness only at low privacy guarantees.
\item LDP obfuscation has, typically, disproporationate impact on only one protected group, and this depends heavily on the outcome $Y$ distribution.
\end{enumerate}

\textbf{Finally, to bridge the gap with practical applications, we frame the observations as concrete recommendations to practitioners considering both ethical concerns of privacy and fairness in ML applications. }

\section{Related Work}
\label{sec:related}

\textbf{Fairness and (L)DP.} To satisfy both privacy and fairness in ML, the literature has proposed several differentially private and fair ML models (e.g., see~\cite{xu2019achieving,Ficiu2023,Jagielski2019,Tran2021} and references within). However, the current state-of-the-art in the intersection field of DP and fairness is multifaceted~\cite{fioretto2022differential}. One perspective aligns DP and fairness in an individual fairness context (e.g.,~\cite{dwork2012fairness}), while the other considers them as opposing forces (e.g.,~\cite{bagdasaryan2019differential,farrand2020neither,ganev2022robin}), considering group fairness. Regarding group fairness notions (our primary focus), the most popular work~\cite{bagdasaryan2019differential} explored the effects of training DP deep learning models, revealing accuracy discrepancies between privileged and unprivileged groups. However, recent studies have begun to observe a negligible~\cite{santana2023empirical} or bounded~\cite{Mangold2023} impact of DP deep learning models on group fairness.
Regarding the local DP setting, some works~\cite{mozannar2020fair,chen2022fairness} proposed to obfuscate only one sensitive attribute under $\epsilon$-LDP guarantees.
With the increasing prevalence of collecting multiple sensitive attributes across various industries, relying solely on LDP for a single sensitive attribute may prove insufficient as correlations can still allow attackers to reconstruct the privatized sensitive attribute. 
For this reason, we consider the case of multiple sensitive attributes, reflecting real-world data collections more accurately.
In this context, a recent work~\cite{Arcolezi2023} has investigated the impact of collecting multi-dimensional under LDP on fairness.
However, while~\cite{Arcolezi2023} has only considered the \textit{independent} setting for randomizing the users multi-dimensional data, for a more comprehensive examination, we have considered both \textit{independent} and \textit{combined} settings (discussed in the following).
Another main difference is that we analyze the impact of LDP on fairness by varying the $Y$ distribution (e.g., see Section~\ref{sub-sec:impact_outcome}).

\textbf{LDP and multi-dimensional data.} Unlike the centralized DP setting, where the server collects users' original data, LDP empowers users to obfuscate their data before transmitting it to the server. While much of the existing literature on LDP has focused on the frequency estimation of one-dimensional data (e.g.,~\cite{kairouz2016discrete,erlingsson2014rappor,apple_2017}), real-world scenarios often involve servers seeking insights into multiple attributes of a population, i.e., \textit{multi-dimensional data}. In this context, the typical process involves local perturbation of user data, followed by statistical estimation and synthetic data generation. The first phase takes place on the user side, while subsequent phases occur at the aggregator/server side. This paper diverges from this typical process in two key aspects. First, it exclusively examines the first phase, studying various approaches for perturbing user multi-dimensional data. Second, it investigates the setup of training an ML model based on the randomized data, as in~\cite{mozannar2020fair,Arcolezi2023,chen2022fairness}. Crucially, the objective is to analyze the impact of different data perturbation approaches on the fairness of the learned model.
In the LDP literature for multi-dimensional data, prior works have adopted either an \textit{independent}~\cite{kikuchi2022castell,lopub}, \textit{sampling-based}~\cite{Filho2023,Arcolezi2021_rs_fd}, or \textit{combined}~\cite{Liu2023} (i.e., joint) perturbation of sensitive attributes. In the former, the randomization mechanism is independently applied to each sensitive attribute, leading to the loss of potentially significant dependencies among attributes and resulting in poor statistical utility. Alternatively, the latter setting, namely combined, treats the Cartesian product of the set of sensitive attributes as a single attribute~\cite{kikuchi2022castell,Liu2023,Domingo2022}, representing a natural approach to sensitive data perturbation. Our goal is then to study the impact of both the independent and combined approaches of user multi-dimensional data perturbation on the fairness of the obtained model. Notice that the sampling-based approach is not comparable since each user only sends information about one attribute. 

\section{Preliminaries and Notation}
\label{sec:preliminaries}

Variables are denoted by capital letters and small letters denote specific values of variables (e.g., $A=a$, $Y=y$).
Bold capital and small letters denote a set of variables and a set of values, respectively.
In particular, $\mathbf{V}$ denotes the set of all variables in the data except the outcome.
A predictor $\hat{Y}$ of an outcome $Y$ is a function of $\mathbf{V}$ ($\hat{Y} = h(\mathbf{V}$)). The set of attributes\footnote{In the rest of the paper we use attribute and variable interchangeably.} $\mathbf{V}$ is composed of non-sensitive ($\mathbf{X}$) and sensitive ($\mathbf{A}$) attributes ($\mathbf{V} = (\mathbf{X},\mathbf{A})$). A \emph{sensitive} attribute reveals private information about an individual and hence should be obfuscated. A \emph{protected} attribute $A$ is an attribute that can be used for discrimination. For example, when deciding to grant a loan to an individual, the protected attribute $A$ could be someone's race or gender. In this paper, we assume that there is only one protected attribute (no intersectionality~\cite{makhlouf2021Bridging}) and the protected attribute is always sensitive ($A\in \mathbf{A}$). 
Note that $\textbf{X}$ could include proxies to $A$ such as zip code which could infer race. Without loss of generality, assume that $\hat{Y}$ and $Y$ are binary random variables where $Y = 1$ (e.g., granting a loan) designates a positive outcome and $Y = 0$ (e.g., denying a loan) designates a negative outcome. In some scenarios, the outcome $Y$ is derived from a score $s$ (e.g., risk score to default on a loan) and a threshold set by domain experts is used to define the cut-off point between the positive outcome and the negative outcome. For the example of granting a loan, an applicant who has a score $s >$ threshold is assigned a positive outcome ($Y = 1$) while an applicant with a score $s <=$ threshold is assigned a negative outcome ($Y = 0$). Thus, varying this threshold causes a variation in the class distribution and potentially leads to different predictions. For the remainder of this paper, we assume that we have access to a dataset $D$ of $n$ i.i.d samples such that $D = {(\mathbf{x}_i,\mathbf{a}_i,y_i)}_{i=1}^{n}$. 
Let $\mathcal{L}$ be a randomization\footnote{The terms randomization and obfuscation are used interchangeably.} algorithm for sensitive attributes. We denote a randomized version of $D$ as $D_z= {(\mathbf{x}_i,\mathbf{z}_i,y_i)}_{i=1}^{n}$ where $\mathbf{z}_i = \mathcal{L}(\mathbf{a}_i)$.

\subsection{Problem Statement}
\label{subsec:problem_statement}
\begin{figure}[!ht]
\centering
    \includegraphics[scale=0.5]{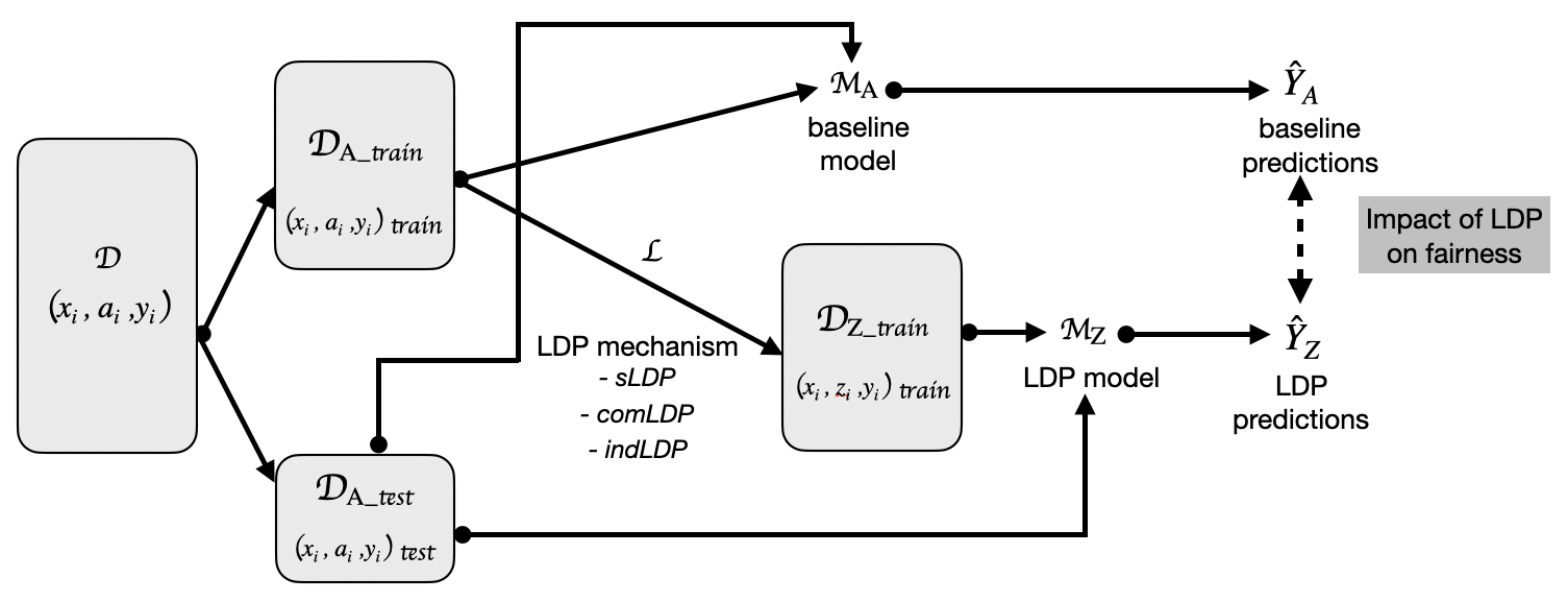} 
    \caption{Our framework for fairness assessment when learning over LDP-based data.}
    \label{fig:framework}
\end{figure}

The focus of this work is to shed light on the impact of training a classifier using locally differential private data on fairness assessment. Figure~\ref{fig:framework} depicts the framework used in this work. To assess fairness, a prediction problem is defined. An example of a prediction problem might be granting loans to individuals or admitting applicants to a college program. As a randomized mechanism $\mathcal{L}$ is applied to the sensitive part of the original data, the predictor $\hat{Y}_Z$ incurs some error.
The difference between $\hat{Y}_Z$ and $\hat{Y}_A$ (predictions of the model trained on the original data) quantifies the impact of LDP on Fairness results. As shown in Figure~\ref{fig:framework}, the classification model called $\mathcal{M}_A$ is first trained using the original data $D_{A_{train}}$. We refer to such a model as a baseline model. We then train the same model with the same hyper-parameters using an obfuscated version of the training set $D_{Z_{train}}$. We call this LDP model $\mathcal{M}_Z$. Note that the classification models $\mathcal{M}_A$ and $\mathcal{M}_Z$ are both tested on the original testing samples ($D_{A_{test}}$).

\subsection{Local Differential Privacy}
\label{subsec:privacy}
This work assumes that the centralized server in charge of aggregating data from individual users is not guaranteed trustworthy. Consequently, we consider the local DP~\cite{kasiviswanathan2011can} setting, where users obfuscate their data before sending it to the server to train the classification model. 

\begin{definition}[$\epsilon$-Local Differential Privacy]\label{def:ldp} An algorithm $\mathcal{L}$ with domain and range equal to the domain of $\mathbf{A}$ ($dom(\mathcal{L}) = range(\mathcal{L}) = dom(\mathbf{A})$) satisfies $\epsilon$-local-differential-privacy ($\epsilon$-LDP), where $\epsilon>0$, if:

\begin{equation*} \label{eq:ldp}
    \max_{\mathbf{a},\mathbf{a'},\mathbf{z} \in dom(\mathbf{A})} \frac{\pr(\mathcal{L}(\mathbf{a}) = \mathbf{z})}{\pr(\mathcal{L}(\mathbf{a'}) = \mathbf{z})} \leq e^\epsilon   
\end{equation*}
\end{definition}

Notice that Definition~\ref{eq:ldp} uses sets of values ($\mathbf{a},\mathbf{a'},$ and $\mathbf{z}$) instead of single values ($a, a',$ and $z$) so that it holds when randomizing one dimensional data (one single sensitive attribute) or multi-dimensional data (several sensitive attributes). The same holds for the $\epsilon$-LDP mechanism defined below. 

\begin{definition} \label{def:krr}
\noindent \textbf{k-Ary Randomized Response (\textbf{$k$}-RR)} Let $\mathbf{A} = \{A_1, A_2, \ldots\}$ be a set of sensitive attributes with a domain $dom(\mathbf{A})=\{\mathbf{a_1},\ldots,\mathbf{a_k}\}$ of size $k$ ($k=|dom(\mathbf{A})|$). Given a value $\mathbf{a} \in dom(\mathbf{A})$, \krr$(\mathbf{a})$ outputs the true value $\mathbf{a}$ with probability $p$, and any other value $\mathbf{a'} \in dom(\mathbf{A}) \setminus \{\mathbf{a}\}$, otherwise. More formally:
\begin{equation} \label{eq:krr}
    \forall{z \in dom(\mathbf{A})}  : \quad \pr(\mathbf{z}=\mathbf{a}) = \begin{cases} p=\frac{e^{\epsilon}}{e^{\epsilon}+k-1} \textrm{ if } \mathbf{z} = \mathbf{a} \textrm{,}\\ q=\frac{1}{e^{\epsilon}+k-1} \textrm{ if } \mathbf{z} \neq \mathbf{a} \textrm{.} \end{cases}
\end{equation}
\noindent where $\mathbf{z}$ is the obfuscated version of $\mathbf{a}$ sent to the server. 
\end{definition}

It is easy to see that $k$-RR mechanism satisfies $\epsilon$-LDP as $\frac{p}{q}=e^{\epsilon}$~\cite{kairouz2016discrete}. As mentioned in Section~\ref{intro}, we choose \krr{} as the LDP mechanism to apply because it does not use any specific user-side encoding, resulting in low computational and communication costs on the user side. Moreover, on the server side, \krr{} does not require any special decoding and has proven optimal for many theoretical information losses in distribution estimation.

\subsection{Fairness}
\label{subsec:fairness}
The common taxonomy of fairness metrics classifies them into group and individual metrics~\cite{10.1145/3457607, 10.1145/3468507.3468511, verma2018fairness,barocas-hardt-narayanan,makhlouf2021Bridging, alves2022survey}. In this paper, we focus on statistical group fairness metrics. These metrics are used to assess the impact of LDP on fairness. 
\begin{itemize}[leftmargin=*]
    \item \textbf{Statistical disparity}\cite{dwork2012fairness} is one of the most commonly applied fairness metrics. It requires the prediction to be statistically independent of the sensitive attribute $(\hat{Y}  \perp A)$. 
In other words, the predicted acceptance rates for both privileged ($A=1$) and unprivileged ($A=0$) groups should be equal. A classifier \^{Y} satisfies statistical parity if:
\begin{equation}
\label{eq:sp}
\mathds{P}(\hat{Y}=1 \mid A = 1) - \mathds{P}(\hat{Y}=1 \mid A = 0).
\end{equation}
    \item \textbf{Equal opportunity disparity}~\cite{hardt2016equality} requires true positive rate equality\footnote{True positive rate = $\frac{TP}{TP+FN}$} among groups:
\begin{equation}
\label{eq:eqOpp}
\mathds{P}(\hat{Y}=1 \mid Y=1,A = 1) - \mathds{P}(\hat{Y}=1\mid Y=1,A = 0).
\end{equation}
    \item \textbf{Predictive equality disparity}~\cite{corbett2017algorithmic} requires only the false positive rates\footnote{False positive rate = $\frac{FP}{FP+TN}$} to be equal in both groups:
\begin{equation}
\label{eq:predEq}
\mathds{P}(\hat{Y}=1 \mid Y=0,A = 1) - \mathds{P}(\hat{Y}=1\mid Y=0,A = 0). 
\end{equation}
    \item \textbf{Overall accuracy disparity}~\cite{berk2021fairness} is satisfied when overall accuracy for both groups is the same:
\begin{equation}
\label{eq:ovAcc}
\mathds{P}(\hat{Y} = Y | A = 1) - \mathds{P}(\hat{Y} = Y | A = 0) 
\end{equation}
    \item \textbf{Predictive rate disparity}~\cite{chouldechova2017fair} requires only the positive predictive value\footnote{Positive predictive value = $\frac{TP}{TP+FP}$} to be equal in both groups and is achieved when:
\begin{equation}
\label{eq:predPar}
\mathds{P}(Y=1 \mid \hat{Y} =1,A = 1) - \mathds{P}(Y=1\mid \hat{Y} =1,A = 0)  
\end{equation}

\end{itemize}

\section{Combined vs independent \krr{}}
\label{sec:combInd}

As noted above, Definitions~\ref{def:ldp} and~\ref{def:krr} hold for the one-dimensional as well as multi-dimensional data. That is, in addition to obfuscating a single sensitive attribute, we also consider the obfuscation of multiple sensitive attributes (Section~\ref{sec:results} presents all the \krr{} settings we consider in this study). More specifically, we assume there are $d$ sensitive attributes $A_1, A_2,\ldots, A_d$, where the domain of each $A_i$ is a discrete set of finite size $k_i=|dom(A_i)|$. We consider two methods to apply \krr{} on multi-dimensional data~\cite{kikuchi2022castell,Domingo2022}:

\begin{description}
\item[Independent \krr]  (\krr-Ind). This is a naive approach that applies \krr{} independently on each attribute. More precisely, \krr-Ind splits the privacy budget $\epsilon$ among the $d$ sensitive attributes, and reports each attribute $A_i$ using $k_i$-RR parameterized with $\epsilon_i$-LDP, where $\sum_i^d\epsilon_i=\epsilon$. 
The state-of-the-art approach~\cite{Arcolezi2021_rs_fd,arcolezi2022improving,kikuchi2022castell,lopub} divides $\epsilon$ evenly among the attributes, i.e., a \textit{uniform} solution in which each attribute is reported under $\frac{\epsilon}{d}$-LDP.
In this study, we apply the \textit{k-based} solution~\cite{Arcolezi2023}. This approach consists of splitting $\epsilon$ among sensitive attributes based on their domain size. More specifically, each sensitive attribute $A_i\in \textbf{A}$ is obfuscated with $\epsilon_{i} = \frac{\epsilon \cdot k_i}{\sum_{i=1}^{d_{s}} k_i}$ where $k_i$ is the domain size of the attribute $A_i$ ($k_i = |dom(A_i)|$) and $d_s$ is the number of attributes in $\textbf{A}$ ($d_s=|\textbf{A}|$).\\

\item[Combined \krr] (\krr-Comb). This mechanism considers the Cartesian product $A_1 \times A_2 \times \ldots \times A_d$ as a single attribute and sanitizes it using \krr{} parameterized with $\epsilon$-LDP, where 
$k=k_1\cdot k_2\cdot\ldots\cdot k_d$. \\
\end{description}

Independent LDP on multi-dimensional data has been studied relatively well in the literature~\cite{lopub,Arcolezi2021_rs_fd,kikuchi2022castell}. Moreover its impact on fairness was the topic of a recent paper ~\cite{Arcolezi2023}. Combined LDP, on the other hand, was not studied extensively. In particular, its impact on fairness is still unclear.

\section{Empirical Results and Analysis}
\label{sec:results}

To study the impact of \krr{} on fairness, two synthetic datasets and two real-world fairness benchmark datasets, namely: \textit{Adult} and \textit{Compas} are used. For each of these datasets, the fairness metrics presented in Section~\ref{subsec:fairness} are applied.

\noindent \textbf{\textit{Environment:}} All the experiments are implemented in Python 3. We use \textit{Random Forest} model~\cite{breiman2001random} for classification with its default hyper-parameters and we use the ten-fold cross-validation technique, both from Scikit-learn~\cite{scikit_learn}. For \krr{} mechanism, we use the implementation in Multi-Freq-LDPy~\cite{arcolezi2022multi}. The codes and datasets for all the experiments are available in the repository~\cite{artifact_impact_ldp_fairness}.

\noindent \textbf{\textit{Stability:}} Since LDP protocols, k-fold cross-validation, and ML algorithms are randomized, we report average results over $20$ runs.

\noindent \textbf{\textit{Datasets:}} A summary of all datasets used in this study is provided in Table~\ref{tab:datasetsInfo}.
\begin{table}[ht]
\footnotesize
\centering
\caption{Metadata of the datasets used in the experiments.}
\label{tab:datasetsInfo}
\renewcommand\arraystretch{1.2}
\begin{tabular}{@{} *6l @{}}    \toprule
\emph{Dataset} & n & A & \textbf{A} & Y & \emph{Threshold} \\
&  &  \emph{(protected att.)} & \emph{(sensitive  att.)} &  \\\midrule
Synthetic & $100$K & A  &- A & Y &$\tau_{Q1}= .44$\\
 &  &   & - C  & & $\tau_{Q2}=.52$\\
 &  &   & - M  & & $\tau_{Q3}=.6$ \\
Compas & $5915$ & race & - race & risk score\footnote{Unlike the synthetic and the \textit{Adult} datasets, whose outcome is continuous, the outcome of the \textit{Compas} dataset is discrete (score $\in [0,1]$). Thus, we use scores 1, 3, and 5 as thresholds for the Y distribution to be skewed to 0, balanced and skewed to 1, respectively.} & $\tau_{Q1}=1$\\
 &  &   &- gender &  & $\tau_{Q2}=3$ \\
 &  &   &- age  &  & $\tau_{Q3}=5$ \\
Adult & $32561$ & gender& - gender  & income  & $\tau_{Q1}=10K$\\
 &  &  &- age   &  & $\tau_{Q2}=27K$\\ 
 &  &   &- race  &  & $\tau_{Q3}=50K$\\ 
  &  &   &- marital-status  & & \\ 
    &  &   &- native-country  & & \\ 
\bottomrule
\hline
\end{tabular}
\end{table}

\begin{figure}[htb]
\begin{minipage}{0.45\textwidth}
\centering
\begin{tikzpicture}[>=latex',scale=0.8]

    \node (C) at (2,0) [] {{$C$}};
    \node (A) at (0,-1) [] {$A$};
    \node (Y) at (4,-1) [] {$Y$};
    \node (M) at (2,-2) [] {$M$};

    \path[every node/.style={sloped,anchor=south,auto=false}]
    (C) edge [->] node {} (A)    
    (C) edge [->] node {} (Y)    
    (A) edge [->] node {} (Y)    
    (A) edge [->] node {} (M)    
    (M) edge [->] node {} (Y);
\end{tikzpicture}
\\ (a) Causal Graph
\label{fig:CG} 
\end{minipage}
\begin{minipage}{0.5\textwidth}
\centering
\begin{equation}\label{eA1}
\begin{split}
C & \sim \mathcal{B}(0.35),\nonumber\\
A & \sim \begin{cases} \mathcal{B}(0.55) \textrm{ if } C=0 \textrm{,}\\ 
    \mathcal{B}(0.75) \textrm{ if } C=1\textrm{,} \end{cases} \nonumber \\
M & \sim \begin{cases} \mathcal{M}(0.35,0.4,0.25) \textrm{ if } A=0 \textrm{,}\\ 
     \mathcal{M}(0.5,0.4,0.1) \textrm{ if } A=1\textrm{,} \end{cases} \nonumber \\
Y & = \alpha A + \beta M + \gamma C + \mathcal{U}_y \textrm{,}\quad \mathcal{U}_y \sim \mathcal{N}(0,1)\nonumber 
\end{split}
\end{equation}
\\ (b) Structural equations
\label{fig:SE} 
\end{minipage}
\caption{Causal Model of the Synthetic Dataset.}
\label{fig:synthetic_model_ACM}
\end{figure}
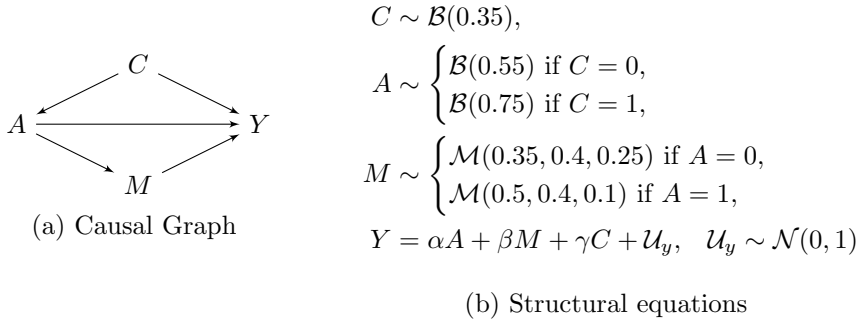

\begin{itemize}
\item[-] \textbf{Synthetic Dataset}: The causal model used to generate the synthetic dataset is depicted in Figure~\ref{fig:synthetic_model_ACM}. $A$, $C$, and $M$ are discrete variables\footnote{C and A follow \textit{Binomial} distributions while M follows \textit{Multinomial} distribution.}, while $Y$ is a continuous variable that is a function of all the other variables such that: $Y= h(A,C,M)$. 
To study the impact of \krr{} on fairness while varying the class distribution, three thresholds are set for the outcome variable $Y$ binarisation, resulting in  
three synthetic datasets differing solely by the distribution of $Y$. The thresholds and the resulting $Y$ distribution for all datasets are shown in Table~\ref{tab:datasetsInfo}. Three scenarios are considered depending on the dataset, namely, $Y$ distribution skewed to $1$, balanced $Y$ distribution, and $Y$ distribution skewed to $0$. 
\item[-] \textbf{Benchmark Datasets}: 
    \subitem - \textit{Compas}: The \textit{Compas} dataset includes data about defendants from Broward County, Florida, during $2013$ and $2014$ who were subject to \textit{Compas} screening. Various information related to the defendants (e.g., race, gender, arrest date, etc.,) were gathered by ProPublica~\cite{angwin2016machine} and the goal is to predict the
two-year violent recidivism. Only black and white defendants assigned \textit{Compas} risk scores within $30$ days of their arrest are kept for analysis leading to $5915$ individuals in total. We consider race as the protected attribute. Five attributes are used in this study namely: race, sex, age, priors and risk score. We use the \textit{Compas} risk score as the outcome. The risk score consists of rating of $1-10$ where the higher the score, the more likely the defendant is to re-offend. Following the same reasoning as the other datasets, we transform the risk score into a binary variable by choosing different thresholds to study the impact of outcome distribution on the privacy-fairness trade-off. Three thresholds are used, leading to three different outcome distributions: skewed to $1$, almost balanced, and skewed to $0$. 
\subitem - \textit{Adult}: The \textit{Adult} dataset\cite{ding2021retiring} consists of $32,561$ samples and the goal is to predict the income of individuals based on several personal attributes such as gender, age, race, marital status, education, and occupation. The attributes considered in this work are age, gender, native country, education level, marital status, number of working hours per week, and income. We use the income of an individual as the outcome. Similarly to the other datasets, different thresholds are used to separate the positive outcome (high income) from the negative outcome (low income). Three thresholds are used in total, leading to three versions of the \textit{Adult} dataset with skewed income distribution to $1$ (threshold = 10K), balanced income distribution (threshold = 26K), and skewed income distribution to $0$ (threshold = 50K\footnote{The 50K threshold is used in the well-known Adult dataset mostly used in the literature~\cite{Dua:2019}.}).
\end{itemize}

\noindent \textbf{\textit{Applied settings:}} 
Four settings (Table~\ref{tab:krr_settings}) are used to assess the impact of LDP on fairness. We vary the privacy level in the range of $\epsilon = \{16,8,5,3,2,1,0.5,0.1\}$. %
\begin{itemize}
\item [-] \textit{noLDP} (Baseline): the model is trained using the original data (without privacy). 
\item [-] \textit{sLDP}: the model is trained using an obfuscated version of the data where only the protected attribute $A$ is obfuscated using \krr{}. 
\item [-] \textit{combLDP}: the model is trained using an obfuscated version of the data where a set of sensitive attributes $\textbf{A}$, including the protected attribute $A$ is obfuscated using \krr-Comb (Section~\ref{sec:combInd}).
\item [-] \textit{indLDP}: the model is trained using an obfuscated version of the data where the same set of sensitive attributes $\textbf{A}$ is obfuscated using \krr-Ind (Section~\ref{subsec:privacy}). The privacy splitting solution used in the experiments is the \textit{k-based} solution (\ref{subsec:privacy}).
\end{itemize}
\begin{table}[ht]
\footnotesize
\centering
\caption{Settings applied in this study.}
\label{tab:krr_settings}
\renewcommand\arraystretch{1.2}
\begin{tabular}{@{} *2l @{}}    \toprule
\emph{Settings applied}  & \emph{k-RR applied to}\\\midrule
$noLDP$ & no privacy \\
$sLDP$ & $A$  \\
$combLDP$ & $\textbf{A}$ using \krr-Comb\\
$indLDP$ & $\textbf{A}$ using \krr-Ind  \\
\bottomrule
\hline
\end{tabular}
\end{table}

\subsection{Impact of LDP on fairness}\label{subsec:ldp_fairness}

This set of experiments aims to study the effect of obfuscating data through LDP on the fairness of the model trained using that data. The experimental protocol consists of obfuscating data using either \textit{sLDP} (not multi-dimensional) or \textit{combLDP} (multi-dimensional) while decreasing the privacy budget $\epsilon$ toward more privacy requirements (small $\epsilon$). Fairness is measured using the various group metrics of Section~\ref{subsec:fairness}, and the experiment is repeated for all three datasets (Synthetic, \textit{Compas}, and \textit{Adult}). Figure~\ref{fig:impact_ldp_fairness} shows the obtained results. To better understand how LDP impacts fairness, the plots show the separate values for both groups: the privileged group ($A=1$) in red dots and the unprivileged group ($A=0$) in blue dots. Disparity between groups is then the difference between the two values (dots). In addition, for a better understanding of the trade-off, disparity in the baseline case (no obfuscation (\textit{noLDP})) is shown using a gray shaded area. The following can be observed from the empirical results.  

\begin{itemize}
    \item \textbf{[Obs1] }\textit{More privacy leads to less disparity.} 
    For both \textit{sLDP} and \textit{combLDP}, the disparity decreases when imposing stronger privacy requirements (smaller $\epsilon$). For example, in Figure~\ref{sub-fig:compas_results_1}, statistical disparity (first row) decreases from $0.23$ to $0.15$ (for \textit{sLDP}) and $0$ (for \textit{combLDP}). The same decreasing pattern can be observed for equal opportunity disparity (second row) and predictive equality disparity (third row). For \textit{overall accuracy disparity}  and \textit{predict rate disparity}, however, disparity either stays unaffected (Figure~\ref{sub-fig:adult_results_1}) or increases (Figures~\ref{sub-fig:syntetic1_results_1} and~\ref{sub-fig:compas_results_1}).
    These two fairness notions compare both groups' accuracy and precision (e.g., $Y=\hat{Y}$ for accuracy). 
    Hence, the behavior is expected since imposing strong privacy guarantees typically leads to a decrease in the accuracy and precision of the classifier for one or both protected groups. But the drop is greater for one group than for the other. This is further detailed when studying the impact of the outcome distribution on the privacy-fairness-utility trade-off (Section~\ref{sub-sec:impact_outcome}).

    \begin{figure}[H]
    \small 
    \begin{subfigure}{0.4\textwidth}
    \includegraphics[scale=0.425]{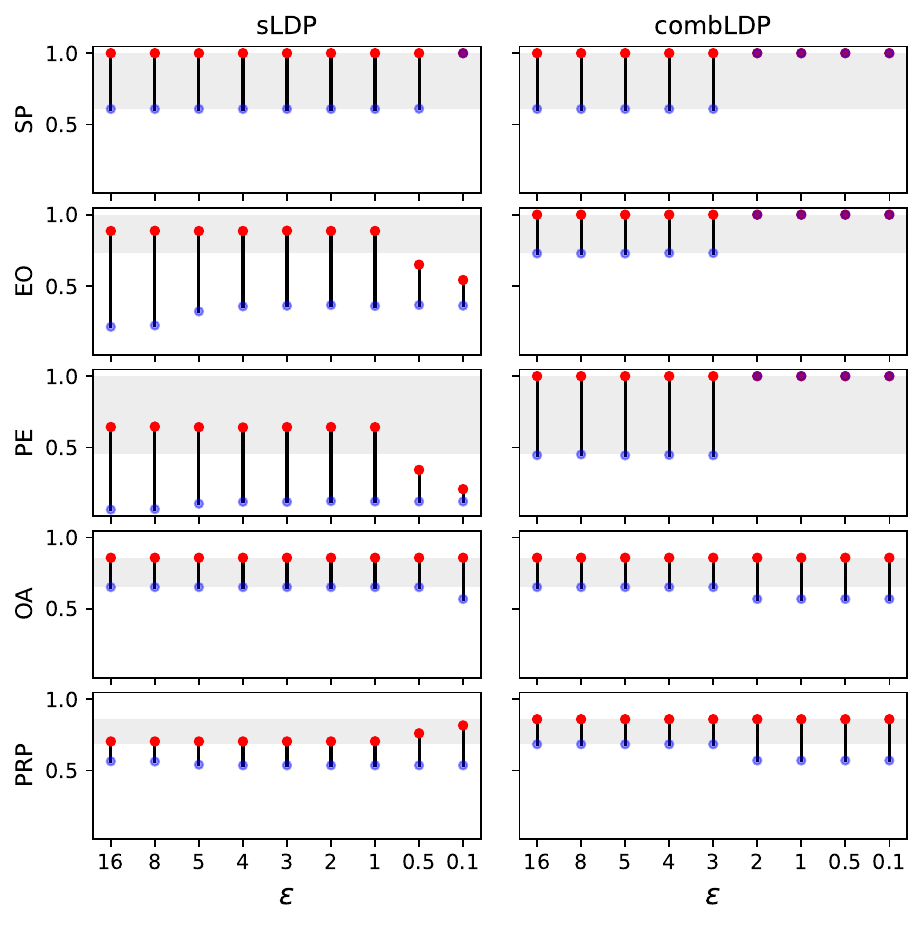} 
    \caption{Synthetic data}
    \label{sub-fig:syntetic1_results_1}
    \end{subfigure} 
    \hspace{20mm}\begin{subfigure}{0.4\textwidth}
    \includegraphics[scale=0.425]{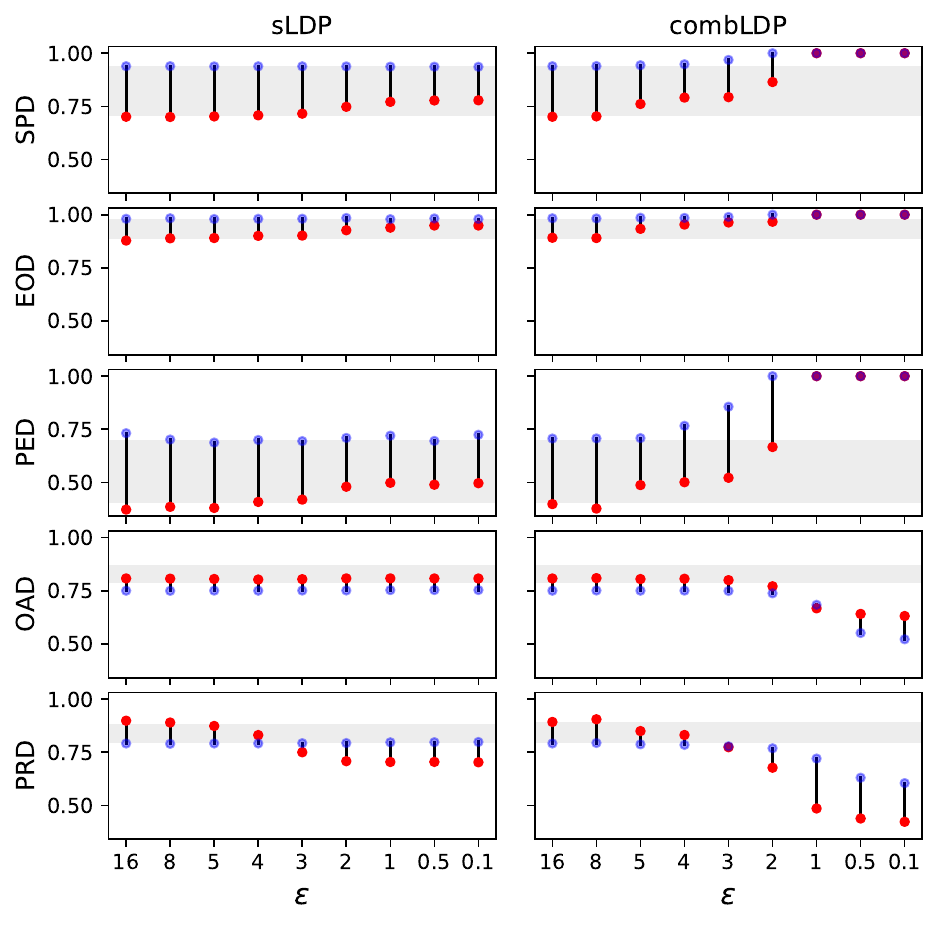} 
    \caption{Compas}
    \label{sub-fig:compas_results_1}
    \end{subfigure}
    \centering
    \begin{subfigure}{0.7\textwidth}
    \includegraphics[scale=0.425]
    {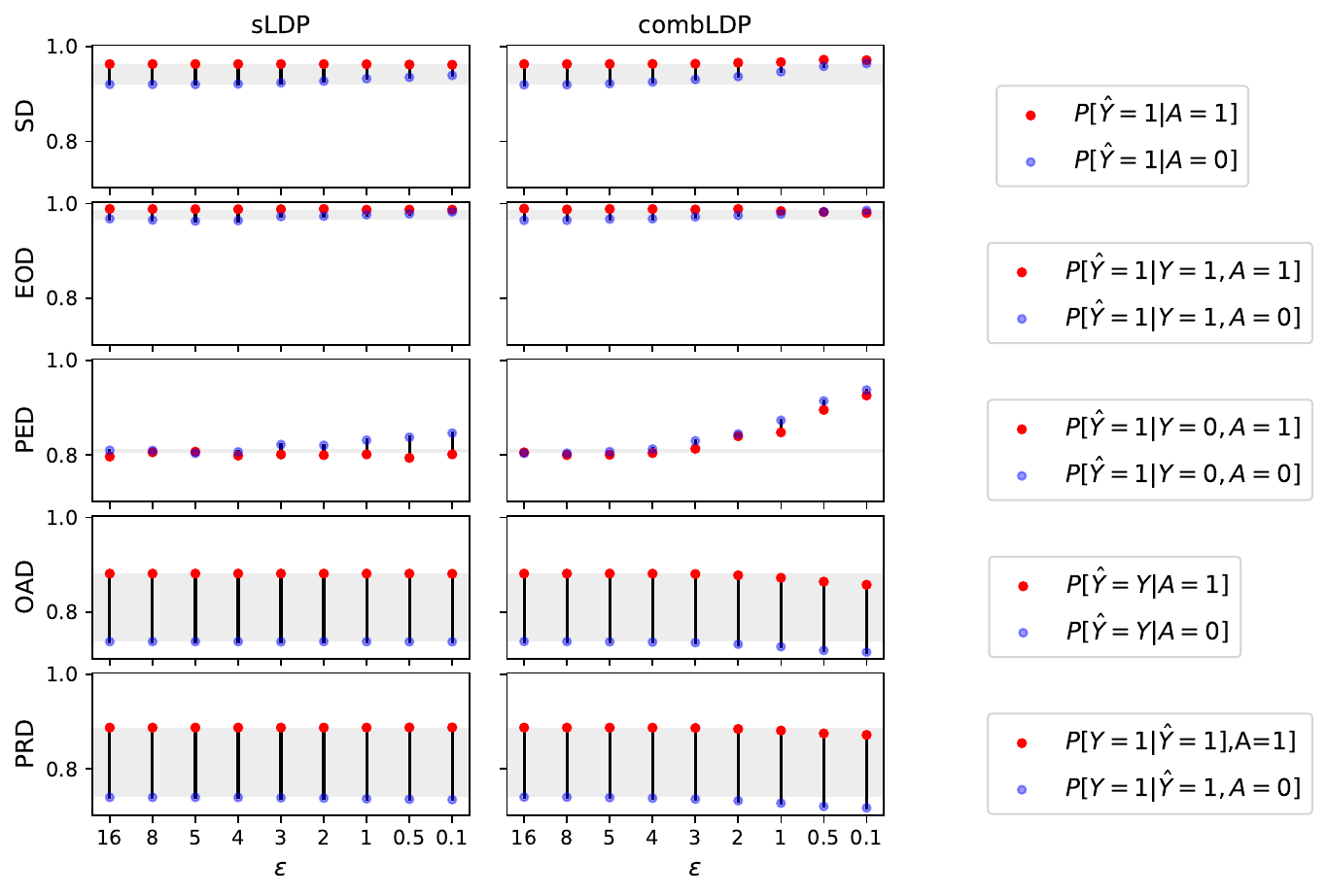} 
    \caption{Adult}
    \label{sub-fig:adult_results_1}
    \end{subfigure}
    \caption{Impact of LDP on disparity (y-axis) by varying the privacy level $\epsilon$ (x-axis). \textit{sLDP} consists in obfuscating a single attribute (protected). \textit{combLDP} consists in obfuscating all sensitive attributes. The gray shaded area represents the disparity results using the baseline model (\textit{noLDP}). 
    }
    \label{fig:impact_ldp_fairness}
\end{figure}
            
    \item \textbf{[Obs2]} \textit{Multi-dimensional LDP reduces disparity more efficiently than one-dimensional LDP.}
    Both \textit{sLDP} and \textit{combLDP} lead to a decrease in disparity (previous observation). However, with \textit{combLDP}, the reduction can be observed with weaker privacy guarantees (higher $\epsilon$). In other words, the more attributes are obfuscated, the less privacy level $\epsilon$ is needed to improve fairness. For instance, in Figure~\ref{sub-fig:syntetic1_results_1}, the disparity disappears at $\epsilon = 0.1$ for \textit{sLDP}, but at $\epsilon = 2$ for \textit{combLDP}. This can be explained by the fact that obfuscating the protected attribute $A$ (equivalent to removing that attribute from the training set when the privacy guarantees are strong enough) is insufficient to improve fairness due to proxies correlated with that attribute. Thus, by additionally obfuscating all attributes correlated with the protected attribute, weaker privacy guarantees are required to reduce the disparity between groups and, therefore, improve fairness.
    
    \item \textbf{[Obs3] } \textit{LDP has disproportionate impact on groups.} In most of the plots, one can observe that \krr{} does not have an impact (or has a minor impact) on one group but a high impact on the other group. For instance, in the first three rows of Figure~\ref{sub-fig:syntetic1_results_1}, the change in disparity is due to a significant change related to only the unprivileged group (blue dots). In other words, considering groups separately, \krr{} impacts the fairness/utility of these groups differently.

\end{itemize}

\subsection{\krr{}-Ind vs \krr{}-Comb} \label{subec:ind-comb}
The impact of LDP on the fairness level of the obtained model depends on the multi-dimensional \krr{} variant (Section~\ref{sec:combInd}) used for obfuscation. The following experiment is performed to compare the effects of \krr{}-Ind and \krr{}-Comb on the disparity between the privileged and unprivileged groups. Benchmark datasets (Synthetic, \textit{Compas}, and \textit{Adult}) are obfuscated using \krr{}-Ind and \krr{}-Comb while decreasing the privacy budget $\epsilon$ toward more strict privacy guarantees (very small $\epsilon$). The obfuscated data is then used to train a predictor and the disparity of the model is then assessed using the fairness metrics of Section~\ref{subsec:fairness}. Figure~\ref{fig:comb-ind} shows the result of the experiments.

\begin{itemize}
    \item \textbf{[Obs4]} \textit{For large $\epsilon$, the efficiency to reduce disparity depends on the sensitive attributes inter-dependencies.} \textit{Compas} and \textit{Adult} experiments illustrate the two different behaviors. In \textit{Compas} experiment (Figure~\ref{sub-fig:compas_results_2}), at $\epsilon = 4$, equal opportunity disparity (EOD) for  \textit{indLDP} is $-0.09$ but $-0.27$ for \textit{combLDP}. Recall, from Table~\ref{tab:datasetsInfo}, that in \textit{Compas} dataset, three attributes are considered sensitive (race, gender, and age) with relatively low inter-dependencies between them. This explains why \krr{}-Ind is more efficient in reducing disparity than \krr{}-Comb for large $\epsilon$ values. In the \textit{Adult} experiment result (Figure~\ref{sub-fig:adult_results_2}), \krr{}-Comb is slightly more efficient than \textit{indLDP} in reducing disparity. For instance, at $\epsilon = 8$, $EOD$ is at $0.43$ for \textit{indLDP} but at $0.38$ for \textit{combLDP}. This can be explained by the relatively high inter-dependencies of the five sensitive attributes (Table~\ref{tab:datasetsInfo}) considered in the \textit{Adult} dataset. 

    \begin{figure}[tb]
        \small
        \begin{subfigure}{0.6\textwidth}
        \includegraphics[scale=0.45]{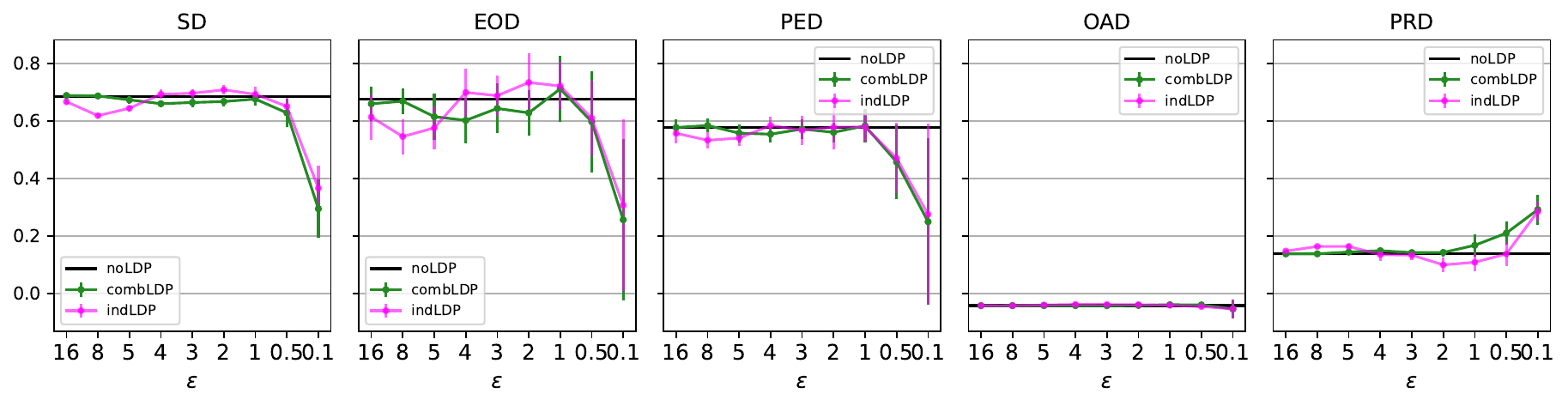} 
        \caption{Synthetic data}
        \label{sub-fig:syntetic1_results_2}
        \end{subfigure}
        \\
        \begin{subfigure}{0.6\textwidth}
        \includegraphics[scale=0.45]{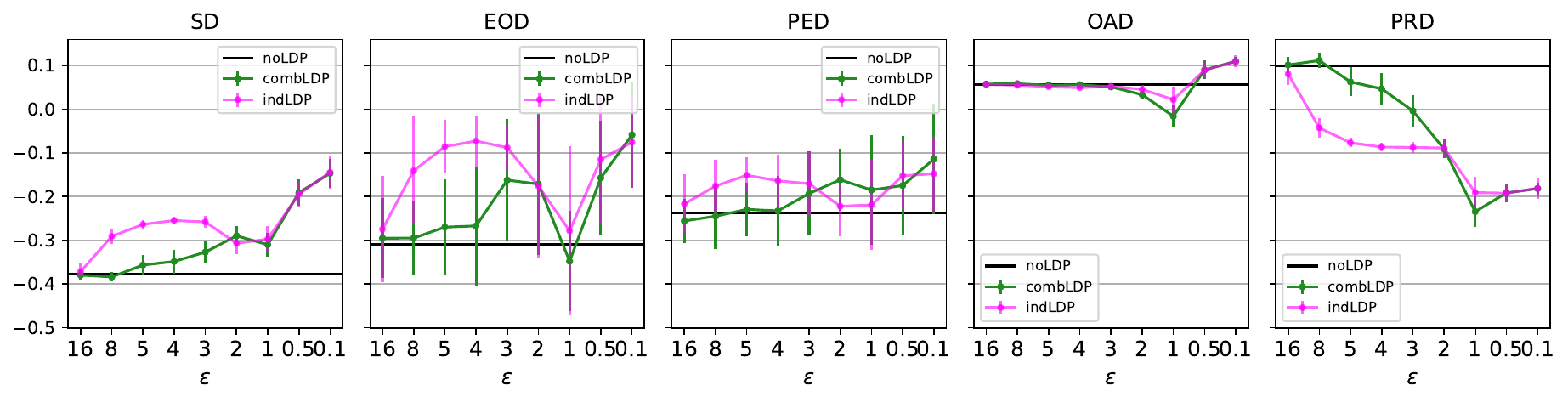} 
        \caption{Compas}
        \label{sub-fig:compas_results_2}
        \end{subfigure}
        \\
        \begin{subfigure}{0.6\textwidth}
        \includegraphics[scale=0.45]
        {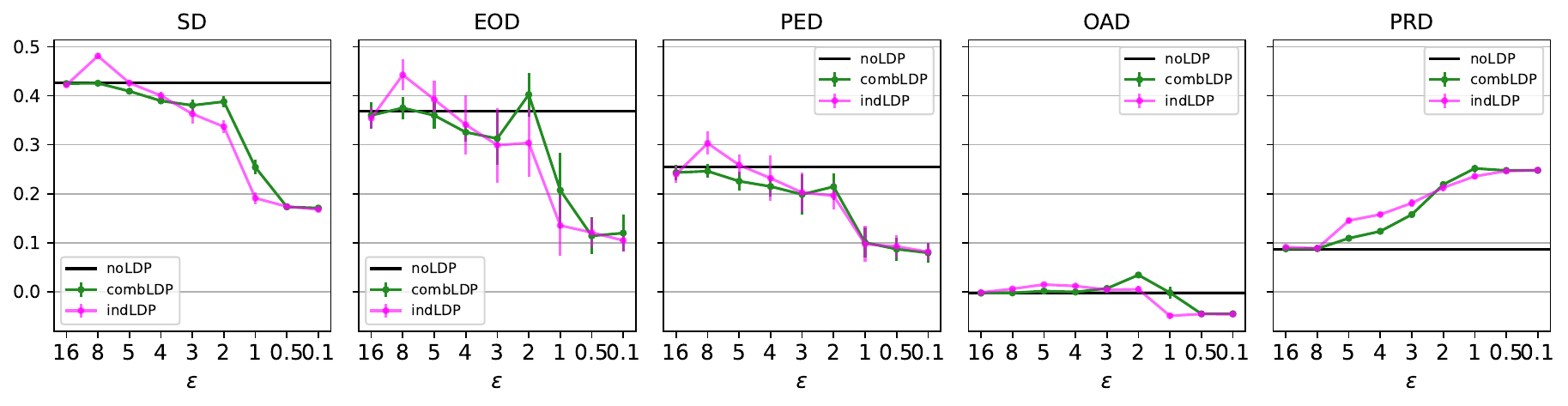} 
        \caption{Adult}
        \label{sub-fig:adult_results_2}
        \end{subfigure}
        \caption{Impact of \textit{combLDP} and \textit{indLDP} on disparity (y-axis) by varying the privacy level $\epsilon$ (x-axis) and obfuscating a set of sensitive attributes. 
        }
        \label{fig:comb-ind}
    \end{figure}
    
    \item \textbf{[Obs5]} \textit{For small $\epsilon$, \textit{combLDP} and \krr{}-Ind have a similar impact on disparity.} In all plots of Figure~\ref{fig:comb-ind}, for strict privacy guarantees (small $\epsilon$), the disparity between protected groups converges to the same value whether the obfuscation was performed with \textit{indLDP} or \textit{indLDP}. In other words, by enforcing more privacy, both settings of \krr{} improved fairness to the same extent.
   
\end{itemize}

\subsection{The effect of changing the outcome distribution}\label{sub-sec:impact_outcome}

\begin{figure}[!htb]
    \centering
    \includegraphics[scale=0.45]{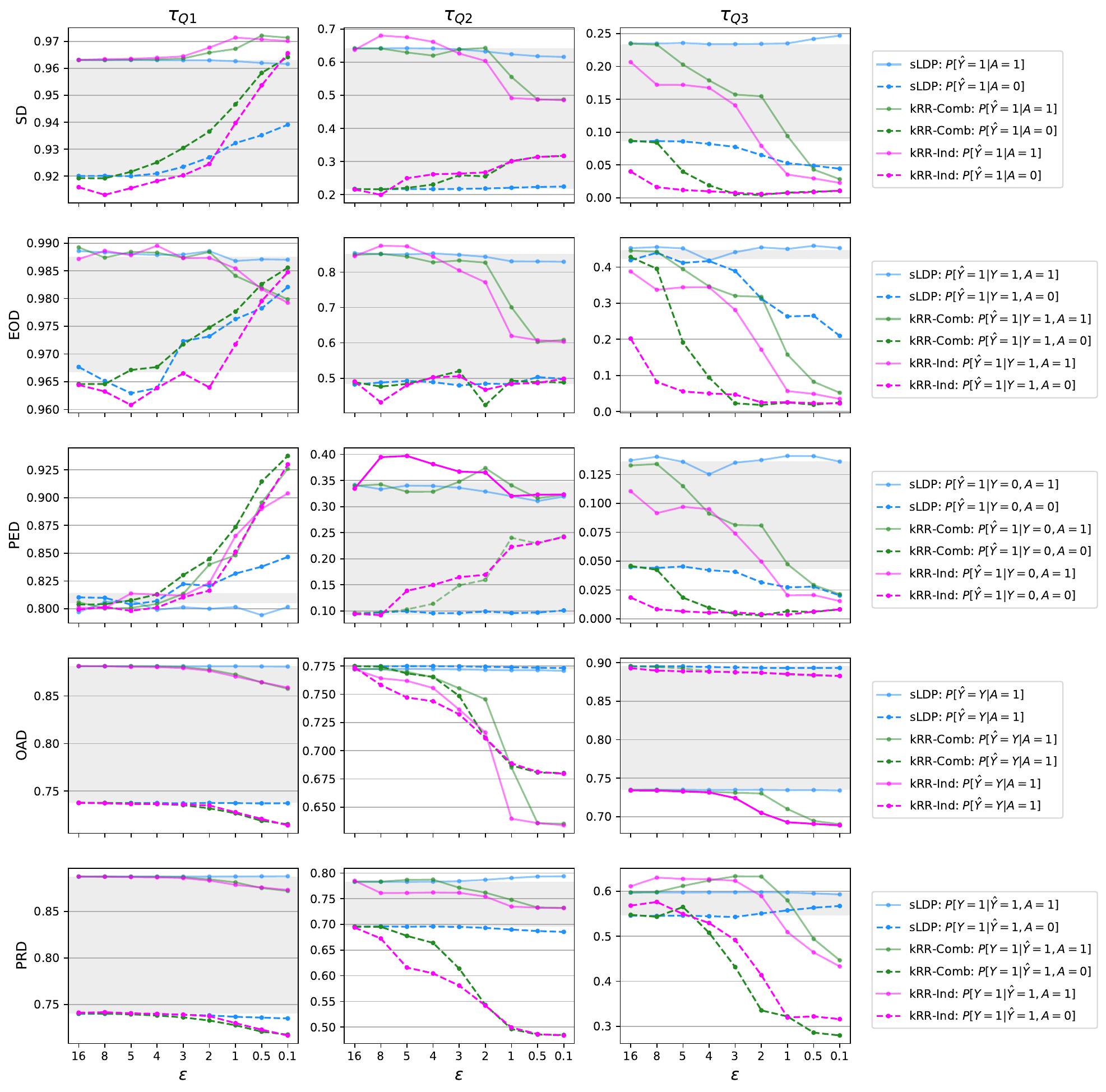} 
    \caption{Impact of Y distribution on the privacy-fairness trade-off. Columns 1, 2, and 3 illustrate the results for the \textit{Adult} dataset when the Y distribution is skewed to 1, balanced, and skewed to 0, respectively.}
    \label{fig:adult_results_3}
\end{figure}

To assess disparity using the group fairness metrics (Section~\ref{subsec:fairness}), the outcome variable $Y$ is required to be binary. However, typically, the trained model predicts a continuous numerical value representing a score as outcome. The score value needs to be thresholded to obtain a binary value. Consequently, the distribution of the outcome variable $Y$ will depend on the threshold value. To study the effect of the outcome distribution on the disparity between protected groups while obfuscating data, the following experiment is performed. Three different distributions are considered for each dataset (Synthetic, \textit{Compas}, and \textit{Adult}). The first distribution is obtained by considering a threshold value ($\tau_{Q1}$) such that all instances in the three top quantiles have positive outcome ($Y=1$). The threshold ($\tau_{Q2}$) of the second distribution is selected such that the two top quantiles have positive outcome. And the third threshold ($\tau_{Q3}$) is selected such that only the instances in the top quantile have positive outcome. Each of the obtained datasets is then obfuscated using \textit{sLDP}, \textit{combLDP}, and \textit{indLDP}. Figure~\ref{fig:adult_results_3} show the experimental results for the Adult dataset (Results for Synthetic and Compas can be found in the appendix (Figures~\ref{fig:syntetic1_results_3} and~\ref{fig:compas_results_3})). As in the experiment of Section~\ref{subsec:ldp_fairness}, to better understand how fairness is impacted by the distribution of the outcome, the plots track the separate values for each protected group (dots on solid lines for privileged group and dots on dashed lines for unprivileged group). The difference between the two types of dots corresponds to the disparity. Finally, as previously mentioned, the grayed area corresponds to the disparity of the baseline model (\textit{noLDP}). 

\begin{itemize}

\item \textbf{[Obs6]} \textit{When enforcing privacy, which group witnesses more accuracy drop depends on the outcome distribution.} Depending on the threshold for positive outcome (and hence the outcome distribution), the drop in accuracy\footnote{As this observation is about the accuracy, only the last two fairness metrics are concerned, that is, OAD and PRD corresponding to the two lower rows of Figure~\ref{fig:adult_results_3}.} due to more tight privacy guarantees (smaller $\epsilon$) is higher for one group than the other. In particular, the accuracy drops more for the unprivileged group $A=0$ when the $Y$ distribution is either skewed to $1$ ($\tau_{Q1}$) or balanced ($\tau_{Q2}$), which correspond to the first and second columns in Figure~\ref{fig:adult_results_3}. Whereas it drops more for the privileged group $A=1$ when the $Y$ distribution is skewed to $0$ ($\tau_{Q1}$)\footnote{Note that this observation is also confirmed in the \textit{Compas} dataset (Figure~\ref{fig:compas_results_3}) but inverted since the privileged group in this dataset is the group $A=0$. To confirm the inversed behavior, we generated a second synthetic dataset where the group $A=0$ is privileged. The plots can be found in Appendix~\ref{subsec:synthetic2}.}.

\item \textbf{[Obs7]} \textit{When enforcing privacy, which group contributes more to reduce the disparity depends on the outcome distribution.} Similarly to the above observation, the outcome distribution has significant impact on how each group (privileged vs unprivileged) contributes to the disparity reduction while enforcing more privacy. In particular, the prediction rates per group (e.g. $P(\hat{Y}=1|A=1)$ for SD) \em{increased} more for the unprivileged group $A=0$ when the outcome distribution is skewed to $1$ ($\tau_{Q1}$ and $\tau_{Q2}$) but \em{decreased} more for the privileged group $A=1$ when the outcome distribution is skewed to $0$ ($\tau_{Q3}$)\footnote{Again, the behavior is reversed for the \textit{Compas} dataset (Figure~\ref{fig:compas_results_3}) for the same reason as the previous observation.}.

\item \textbf{[Obs8]} \textit{For a fair baseline model, enforcing privacy amplifies disparity.} The outcome distribution experiment exhibited an interesting behavior illustrated clearly in the \textit{Adult} dataset results (Figure~\ref{fig:adult_results_3}). In particular, for the PED metric with outcome distribution at threshold $\tau_{Q1}$, the disparity in the baseline predictor is relatively small. However, training the predictor using obfuscated data resulted in disparity amplification. A similar behavior is observed for OAD with $\tau_{Q2}$.  

\end{itemize}

Based on the above observations, one can conclude the following statements:

\noindent \textbf{\textit{Statement 1:}}\textit{
If $A=a$ is the privileged group (has a majority of $Y=1$)  then if $Y$ is skewed to $1$, adding noise affects more the accuracy of the unprivileged group $A\neq a$ 
else ($Y$ is skewed to $0$) adding noise affects more the accuracy of $A = a$.}

\noindent \textbf{\textit{Statement 2:}}
\textit{If $A=a$ is the privileged group (has a majority of $Y=1$), then if $Y$ is skewed to $1$, adding noise increases more the predicted rates for the unprivileged group $A\neq a$ else ($Y$ is skewed to $0$),  adding noise decreases more the predicted rates for group $A = a$.}

\subsection{Recommendations}\label{sec:discussion}

Based on the observations obtained from the experimental analysis, one can propose the following recommendations for a practitioner who is considering a mechanism satisfying privacy and fairness guarantees. That is, a mechanism allowing individual users to share their data while at the same time protecting their sensitive information and guaranteeing that the obtained model is fair with respect to sub-populations and/or individuals. 

\paragraph{A. LDP data obfuscation is an efficient mechanism to reduce disparity.} Almost all observations from the experimental analysis confirm the conclusion that LDP obfuscation reduces disparity (\textbf{Obs1, Obs2, Obs4, Obs7}). The disparity reduction is often due to one group being more sensitive to the LDP obfuscation rather the other (\textbf{Obs2}). The only exception is when the predicted model using baseline (not obfuscated) data is already fair. In that case, LDP may create disparity (\textbf{Obs8}). 

\paragraph{B. Obfuscating several sensitive attributes allows to reduce disparity more efficiently than a single attribute.}
If a practioner is interested in producing a fair model but with a minimal privacy enforcement, it is recommended that she uses multi-dimensional LDP obfuscating as many sensitive attributes as possible (\textbf{Obs2}).

\paragraph{C. Independent and combined variants of multi-dimensional LDP are different only with weak privacy guarantees.}
The choice of the multi-dimensional approach of LDP (combined vs independent) matters only at low privacy guarantees (large $\epsilon$) (\textbf{Obs4}). In that case, the practitioner's choice should depend on the level of interdependency between sensitive attribute. For high interdependency, a combined approach is more efficient to reduce disparity. For low or no interdependency, an independent approach is more efficient. At strict privacy guarantees (low $\epsilon$), however, both approaches have similar effect on disparity (\textbf{Obs5}).

\paragraph{D. Obfuscating data impacts disproportionally only one group depending on the outcome distribution.}
A practitioner who obfuscates individual data with LDP should expect that only one group will be significantly affected. And she can \textit{guess} which group will be more affected by studying the outcome distribution. More precisely, if the outcome distribution is skewed towards the positive outcome (typically $Y=1$), it is the unpriviliged group who will be more affected. Otherwise (outcome distribution is skewed to the negative outcome (typically $Y=0$), it is the privileged group who will be more affected (\textbf{Obs7} and \textbf{Obs8}).

\section{Conclusion}
\label{sec:conclusion}

This paper investigates how the accuracy and fairness of the decisions made by the model change under local differential privacy (LDP), in particular, $k$-ary Randomized Response ($k$-RR) mechanism, given different levels of privacy and different class distributions. To broaden the scope of our study, we employed various group fairness metrics and evaluated two settings for obfuscating multi-dimensional sensitive attributes under LDP, namely, independent and combined, on one synthetic and two benchmark datasets to substantiate our claims.
The experimental analysis revealed very relevant observations that we framed as concrete recommendations for machine learning practitioners aiming at guaranteeing both ethical concerns of privacy and fairness. 
To the best of our knowledge, this is the first work which studies the effect of combined multi-dimensional LDP on fairness. In particular, we observed that combined LDP reduces more efficiently the disparity at low privacy guarantees (high $\epsilon$). 
As future work, we aim to formalize the privacy-utility-fairness trade-off when learning over LDP-based data, as well as to propose LDP- and fairness-aware ML models.

\bibliographystyle{plain}
\bibliography{ref}

\begin{thebibliography}{10}

\bibitem{artifact_impact_ldp_fairness}
Impact ldp on fairness repository.
\newblock \url{https://github.com/KarimaMakhlouf/Impact_of_LDP_on_Fairness}.

\bibitem{alves2022survey}
Guilherme Alves, Fabien Bernier, Miguel Couceiro, Karima Makhlouf, Catuscia Palamidessi, and Sami Zhioua.
\newblock Survey on fairness notions and related tensions.
\newblock {\em arXiv preprint arXiv:2209.13012}, 2022.

\bibitem{angwin2016machine}
Julia Angwin, Jeff Larson, Surya Mattu, and Lauren Kirchner.
\newblock Machine bias. propublica.
\newblock {\em See https://www. propublica. org/article/machine-bias-risk-assessments-in-criminal-sentencing}, 2016.

\bibitem{apple_2017}
Differential Privacy~Team Apple.
\newblock Learning with privacy at scale, Dec 2017.

\bibitem{Arcolezi2021_rs_fd}
H\'{e}ber~H. Arcolezi, Jean-Fran\c{c}ois Couchot, Bechara Al~Bouna, and Xiaokui Xiao.
\newblock Random sampling plus fake data: Multidimensional frequency estimates with local differential privacy.
\newblock In {\em Proceedings of the 30th ACM International Conference on Information \& Knowledge Management}, CIKM '21, page 47–57, New York, NY, USA, 2021. Association for Computing Machinery.

\bibitem{arcolezi2022multi}
H{\'e}ber~H. Arcolezi, Jean-Fran{\c{c}}ois Couchot, S{\'e}bastien Gambs, Catuscia Palamidessi, and Majid Zolfaghari.
\newblock Multi-freq-ldpy: Multiple frequency estimation under local differential privacy in python.
\newblock In Vijayalakshmi Atluri, Roberto Di~Pietro, Christian~D. Jensen, and Weizhi Meng, editors, {\em Computer Security -- ESORICS 2022}, pages 770--775, Cham, 2022. Springer Nature Switzerland.

\bibitem{Arcolezi2023}
H{\'{e}}ber~H. Arcolezi, Karima Makhlouf, and Catuscia Palamidessi.
\newblock (local) differential privacy has {NO} disparate impact on~fairness.
\newblock In {\em Data and Applications Security and Privacy {XXXVII}}, pages 3--21. Springer Nature Switzerland, 2023.

\bibitem{arcolezi2022improving}
Héber~H. Arcolezi, Jean-François Couchot, Bechara {Al Bouna}, and Xiaokui Xiao.
\newblock Improving the utility of locally differentially private protocols for longitudinal and multidimensional frequency estimates.
\newblock {\em Digital Communications and Networks}, 2022.

\bibitem{bagdasaryan2019differential}
Eugene Bagdasaryan, Omid Poursaeed, and Vitaly Shmatikov.
\newblock Differential privacy has disparate impact on model accuracy.
\newblock {\em Advances in neural information processing systems}, 32, 2019.

\bibitem{barocas-hardt-narayanan}
Solon Barocas, Moritz Hardt, and Arvind Narayanan.
\newblock {\em Fairness and Machine Learning}.
\newblock fairmlbook.org, 2019.
\newblock \url{http://www.fairmlbook.org}.

\bibitem{berk2021fairness}
Richard Berk, Hoda Heidari, Shahin Jabbari, Michael Kearns, and Aaron Roth.
\newblock Fairness in criminal justice risk assessments: The state of the art.
\newblock {\em Sociological Methods \& Research}, 50(1):3--44, 2021.

\bibitem{breiman2001random}
Leo Breiman.
\newblock Random forests.
\newblock {\em Machine learning}, 45:5--32, 2001.

\bibitem{chang2021privacy}
Hongyan Chang and Reza Shokri.
\newblock On the privacy risks of algorithmic fairness.
\newblock In {\em 2021 IEEE European Symposium on Security and Privacy (EuroS\&P)}, pages 292--303. IEEE, 2021.

\bibitem{chen2022fairness}
Canyu Chen, Yueqing Liang, Xiongxiao Xu, Shangyu Xie, Yuan Hong, and Kai Shu.
\newblock When fairness meets privacy: Fair classification with semi-private sensitive attributes.
\newblock In {\em Workshop on Trustworthy and Socially Responsible Machine Learning, NeurIPS 2022}, 2022.

\bibitem{chouldechova2017fair}
Alexandra Chouldechova.
\newblock Fair prediction with disparate impact: A study of bias in recidivism prediction instruments.
\newblock {\em Big data}, 5(2):153--163, 2017.

\bibitem{corbett2017algorithmic}
Sam Corbett-Davies, Emma Pierson, Avi Feller, Sharad Goel, and Aziz Huq.
\newblock Algorithmic decision making and the cost of fairness.
\newblock In {\em Proceedings of the 23rd ACM SIGKDD International Conference on Knowledge Discovery and Data Mining}, pages 797--806, 2017.

\bibitem{Filho2023}
José~Serafim Costa~Filho and Javam~C Machado.
\newblock Felip: A local differentially private approach to frequency estimation on multidimensional datasets.
\newblock In {\em Proceedings of the 26th International Conference on Extending Database Technology, {EDBT} 2023, Ioannina, Greece, March 28 - March 31, 2023}, pages 671--683. OpenProceedings.org, 2023.

\bibitem{santana2023empirical}
Anderson~Santana de~Oliveira, Caelin Kaplan, Khawla Mallat, and Tanmay Chakraborty.
\newblock An empirical analysis of fairness notions under differential privacy.
\newblock {\em arXiv preprint arXiv:2302.02910}, 2023.

\bibitem{ding2021retiring}
Frances Ding, Moritz Hardt, John Miller, and Ludwig Schmidt.
\newblock Retiring adult: New datasets for fair machine learning.
\newblock {\em Advances in Neural Information Processing Systems}, 34, 2021.

\bibitem{Domingo2022}
Josep Domingo-Ferrer and Jordi Soria-Comas.
\newblock Multi-dimensional randomized response.
\newblock {\em IEEE Transactions on Knowledge and Data Engineering}, 34(10):4933--4946, 2022.

\bibitem{Dua:2019}
Dheeru Dua and Casey Graff.
\newblock {UCI} machine learning repository, 2017.

\bibitem{dwork2012fairness}
Cynthia Dwork, Moritz Hardt, Toniann Pitassi, Omer Reingold, and Richard Zemel.
\newblock Fairness through awareness.
\newblock In {\em Proceedings of the 3rd innovations in theoretical computer science conference}, pages 214--226, 2012.

\bibitem{dwork2016calibrating}
Cynthia Dwork, Frank McSherry, Kobbi Nissim, and Adam Smith.
\newblock Calibrating noise to sensitivity in private data analysis.
\newblock In {\em Theory of Cryptography}, pages 265--284. Springer Berlin Heidelberg, 2006.

\bibitem{erlingsson2014rappor}
{\'U}lfar Erlingsson, Vasyl Pihur, and Aleksandra Korolova.
\newblock Rappor: Randomized aggregatable privacy-preserving ordinal response.
\newblock In {\em Proceedings of the 2014 ACM SIGSAC conference on computer and communications security}, pages 1054--1067, 2014.

\bibitem{farrand2020neither}
Tom Farrand, Fatemehsadat Mireshghallah, Sahib Singh, and Andrew Trask.
\newblock Neither private nor fair: Impact of data imbalance on utility and fairness in differential privacy.
\newblock In {\em Proceedings of the 2020 Workshop on Privacy-Preserving Machine Learning in Practice}, pages 15--19, 2020.

\bibitem{Ficiu2023}
Bogdan Ficiu, Neil~D. Lawrence, and Andrei Paleyes.
\newblock Automated discovery of trade-off between utility, privacy and fairness in machine learning models.
\newblock {\em arXiv preprint arXiv:2311.15691}, 2023.

\bibitem{fioretto2022differential}
Ferdinando Fioretto, Cuong Tran, Pascal Van~Hentenryck, and Keyu Zhu.
\newblock Differential privacy and fairness in decisions and learning tasks: A survey.
\newblock {\em arXiv preprint arXiv:2202.08187}, 2022.

\bibitem{ganev2022robin}
Georgi Ganev, Bristena Oprisanu, and Emiliano De~Cristofaro.
\newblock Robin hood and matthew effects: Differential privacy has disparate impact on synthetic data.
\newblock In {\em International Conference on Machine Learning}, pages 6944--6959. PMLR, 2022.

\bibitem{hardt2016equality}
Moritz Hardt, Eric Price, and Nati Srebro.
\newblock Equality of opportunity in supervised learning.
\newblock {\em Advances in neural information processing systems}, 29:3315--3323, 2016.

\bibitem{Jagielski2019}
Matthew Jagielski, Michael Kearns, Jieming Mao, Alina Oprea, Aaron Roth, Saeed~Sharifi Malvajerdi, and Jonathan Ullman.
\newblock Differentially private fair learning.
\newblock In Kamalika Chaudhuri and Ruslan Salakhutdinov, editors, {\em Proceedings of the 36th International Conference on Machine Learning}, volume~97 of {\em Proceedings of Machine Learning Research}, pages 3000--3008. PMLR, 09--15 Jun 2019.

\bibitem{kairouz2016discrete}
Peter Kairouz, Keith Bonawitz, and Daniel Ramage.
\newblock Discrete distribution estimation under local privacy.
\newblock In {\em International Conference on Machine Learning}, pages 2436--2444. PMLR, 2016.

\bibitem{kasiviswanathan2011can}
Shiva~Prasad Kasiviswanathan, Homin~K Lee, Kobbi Nissim, Sofya Raskhodnikova, and Adam Smith.
\newblock What can we learn privately?
\newblock {\em SIAM Journal on Computing}, 40(3):793--826, 2011.

\bibitem{kikuchi2022castell}
Hiroaki Kikuchi.
\newblock Castell: Scalable joint probability estimation of multi-dimensional data randomized with local differential privacy.
\newblock {\em arXiv preprint arXiv:2212.01627}, 2022.

\bibitem{Liu2023}
Gaoyuan Liu, Peng Tang, Chengyu Hu, Chongshi Jin, and Shanqing Guo.
\newblock Multi-dimensional data publishing with local differential privacy.
\newblock In {\em Proceedings of the 26th International Conference on Extending Database Technology, {EDBT} 2023, Ioannina, Greece, March 28 - March 31, 2023}, pages 183--194. OpenProceedings.org, 2023.

\bibitem{makhlouf2021Bridging}
Karima Makhlouf, Sami Zhioua, and Catuscia Palamidessi.
\newblock Machine learning fairness notions: Bridging the gap with real-world applications.
\newblock {\em Information Processing \& Management}, 58(5):102642, 2021.

\bibitem{10.1145/3468507.3468511}
Karima Makhlouf, Sami Zhioua, and Catuscia Palamidessi.
\newblock On the applicability of machine learning fairness notions.
\newblock 23(1):14–23, may 2021.

\bibitem{Mangold2023}
Paul Mangold, Micha\"{e}l Perrot, Aur\'{e}lien Bellet, and Marc Tommasi.
\newblock Differential privacy has bounded impact on fairness in classification.
\newblock In Andreas Krause, Emma Brunskill, Kyunghyun Cho, Barbara Engelhardt, Sivan Sabato, and Jonathan Scarlett, editors, {\em Proceedings of the 40th International Conference on Machine Learning}, volume 202 of {\em Proceedings of Machine Learning Research}, pages 23681--23705. PMLR, 23--29 Jul 2023.

\bibitem{10.1145/3457607}
Ninareh Mehrabi, Fred Morstatter, Nripsuta Saxena, Kristina Lerman, and Aram Galstyan.
\newblock A survey on bias and fairness in machine learning.
\newblock {\em ACM Comput. Surv.}, 54(6), jul 2021.

\bibitem{mitchell2021algorithmic}
Shira Mitchell, Eric Potash, Solon Barocas, Alexander D'Amour, and Kristian Lum.
\newblock Algorithmic fairness: Choices, assumptions, and definitions.
\newblock {\em Annual Review of Statistics and Its Application}, 8:141--163, 2021.

\bibitem{mozannar2020fair}
Hussein Mozannar, Mesrob Ohannessian, and Nathan Srebro.
\newblock Fair learning with private demographic data.
\newblock In {\em International Conference on Machine Learning}, pages 7066--7075. PMLR, 2020.

\bibitem{scikit_learn}
F.~Pedregosa, G.~Varoquaux, A.~Gramfort, V.~Michel, B.~Thirion, O.~Grisel, M.~Blondel, P.~Prettenhofer, R.~Weiss, V.~Dubourg, J.~Vanderplas, A.~Passos, D.~Cournapeau, M.~Brucher, M.~Perrot, and E.~Duchesnay.
\newblock Scikit-learn: Machine learning in {P}ython.
\newblock {\em Journal of Machine Learning Research}, 12:2825--2830, 2011.

\bibitem{lopub}
Xuebin Ren, Chia-Mu Yu, Weiren Yu, Shusen Yang, Xinyu Yang, Julie~A McCann, and S~Yu Philip.
\newblock Lopub: high-dimensional crowdsourced data publication with local differential privacy.
\newblock {\em IEEE Transactions on Information Forensics and Security}, 13(9):2151--2166, 2018.

\bibitem{Tran2021}
Cuong Tran, Ferdinando Fioretto, and Pascal Van~Hentenryck.
\newblock Differentially private and fair deep learning: A lagrangian dual approach.
\newblock {\em Proceedings of the AAAI Conference on Artificial Intelligence}, 35(11):9932–9939, May 2021.

\bibitem{verma2018fairness}
Sahil Verma and Julia Rubin.
\newblock Fairness definitions explained.
\newblock In {\em 2018 IEEE/ACM International Workshop on Software Fairness (FairWare)}, pages 1--7. IEEE, 2018.

\bibitem{xu2019achieving}
Depeng Xu, Shuhan Yuan, and Xintao Wu.
\newblock Achieving differential privacy and fairness in logistic regression.
\newblock In {\em Companion proceedings of The 2019 world wide web conference}, pages 594--599, 2019.

\end{thebibliography}
\newpage
\appendix

\section{Appendix}
\label{sec:appendix}

\subsection{Results of the Synthetic dataset 2}\label{subsec:synthetic2}
The synthetic dataset 2 follows the exact same causal model depicted in Figure~\ref{fig:synthetic_model_ACM}. The data distribution is the only difference between the Synthetic datasets 1 and 2. More specifically, synthetic data 2 differed from synthetic data 1 solely by Y distribution.

\begin{figure}[H]
    \centering
    \includegraphics[scale=0.4]{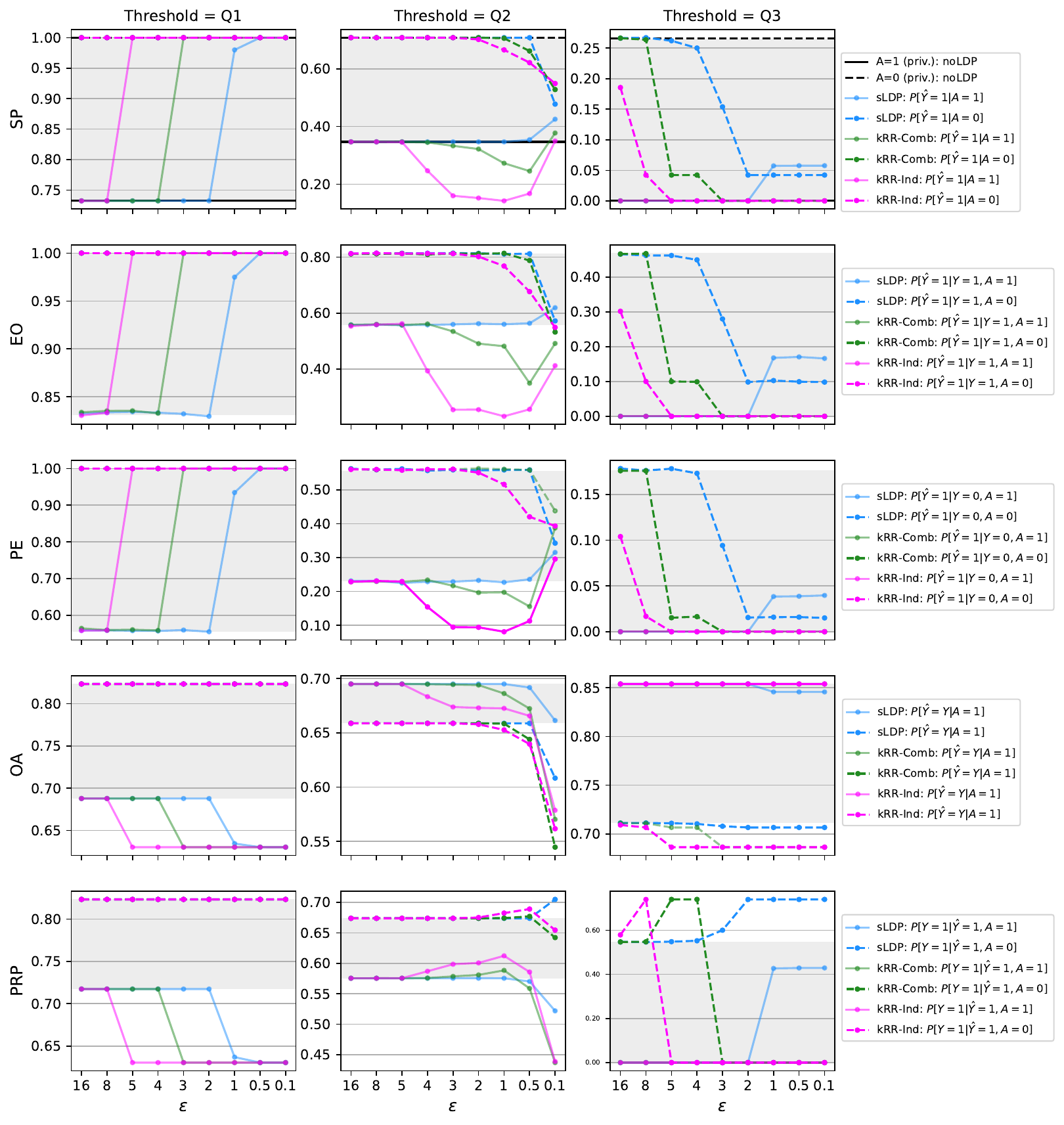} 
    \caption{Impact of \krr{} on fairness for the Adult datasets generated with three different thresholds leading to different Y distributions. Synthetic data 2}
    \label{fig:synthetic2_results_3}
\end{figure}

\subsection{Synthetic and Compas experimental results for Section~\ref{sub-sec:impact_outcome}}

\begin{figure}[H]
    \centering
    \includegraphics[scale=0.4]{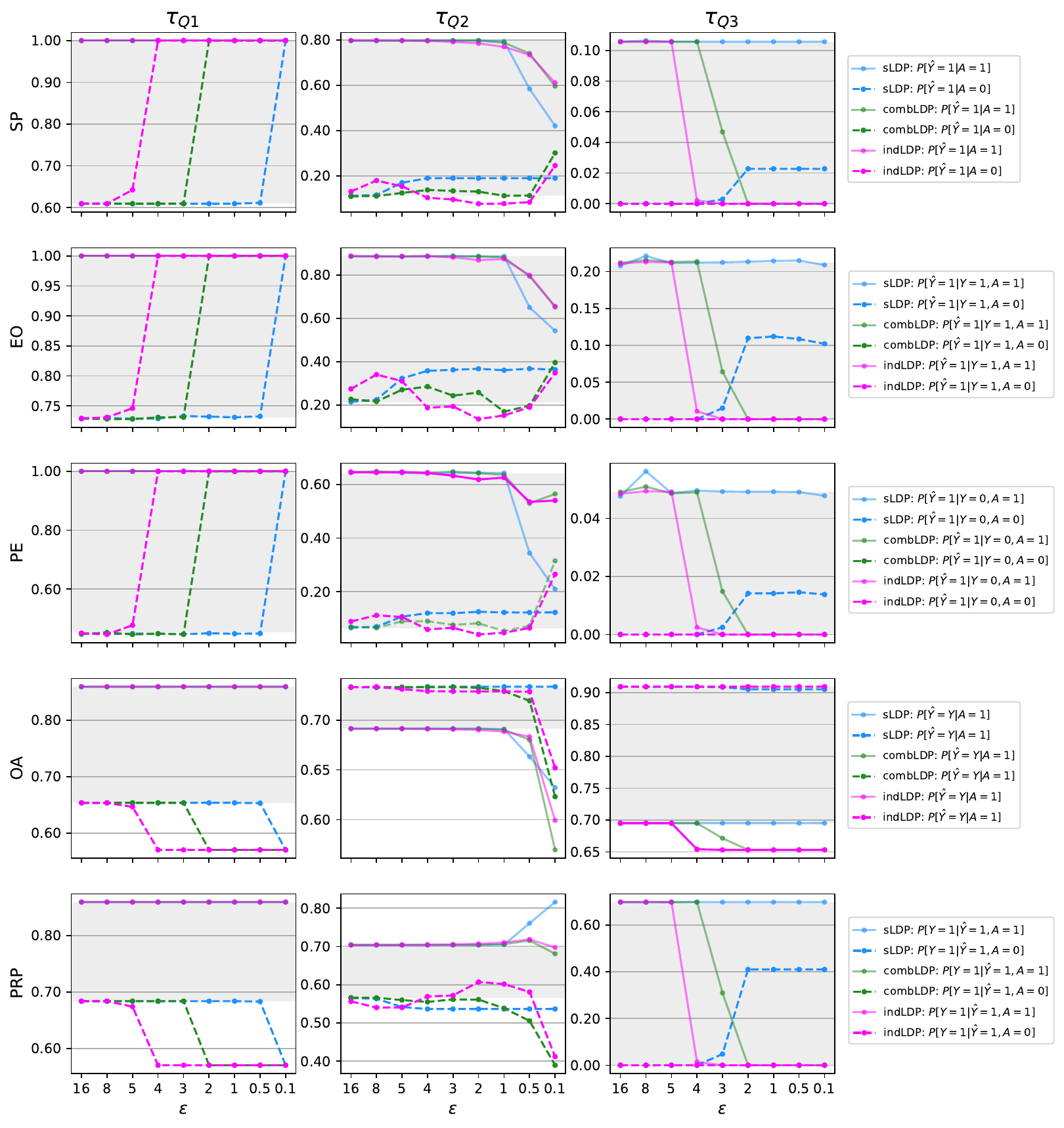} 
    \caption{Impact of Y distribution on the privacy-fairness trade-off. Columns 1, 2, and 3 illustrate the results for the synthetic dataset when the Y distribution is skewed to 1, balanced, and skewed to 0, respectively.}
    \label{fig:syntetic1_results_3}
\end{figure}

\begin{figure}[H]
    \centering
    \includegraphics[scale=0.4]{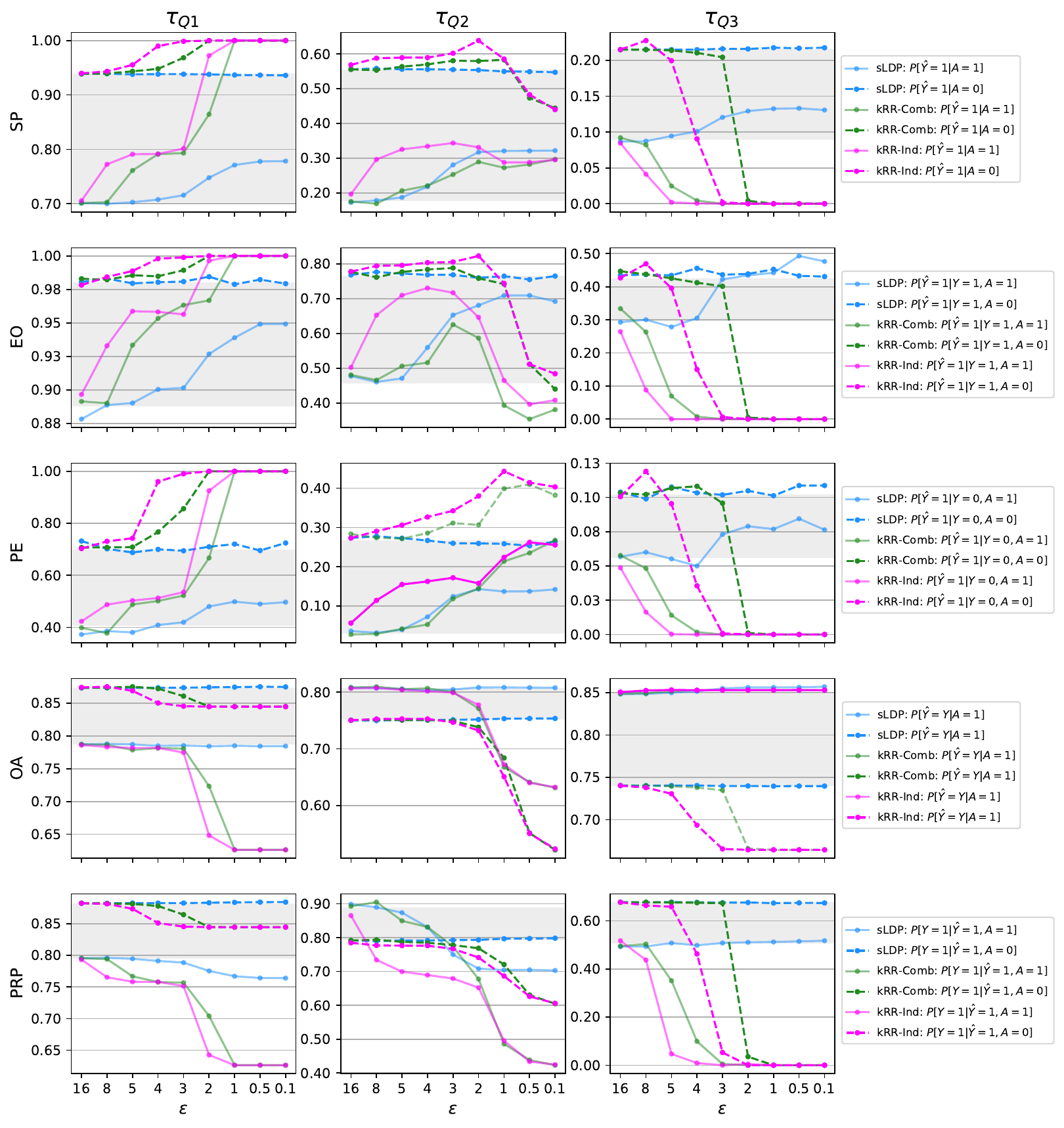} 
    \caption{Impact of Y distribution on the privacy-fairness trade-off. Columns 1, 2, and 3 illustrate the results for the \textit{Compas} dataset when the Y distribution is skewed to 1, balanced, and skewed to 0, respectively.}
    \label{fig:compas_results_3}
\end{figure}

\end{document}